\documentclass[letterpaper]{article} 
\usepackage{CameraReady/LaTeX/aaai2026}
\usepackage{times}  
\usepackage{helvet}  
\usepackage{courier}  
\usepackage[hyphens]{url}  
\usepackage{graphicx} 
\usepackage{natbib}  
\usepackage{caption} 
\usepackage{algorithm}
\usepackage{algorithmic}
\usepackage{amsmath, amssymb, amsthm}
\usepackage{multirow}
\usepackage{booktabs}
\usepackage{array}
\usepackage[table]{xcolor}
\usepackage{dsfont}
\usepackage{subcaption}
\usepackage{hyperref}
\usepackage[most]{tcolorbox}

\usepackage{newfloat}
\usepackage{listings}
\floatstyle{ruled}
\newfloat{listing}{tb}{lst}{}
\floatname{listing}{Listing}

\newcolumntype{C}[1]{>{\centering\arraybackslash}m{#1}}
\DeclareCaptionStyle{ruled}{labelfont=normalfont,labelsep=colon,strut=off} 
\floatstyle{ruled}
\newfloat{listing}{tb}{lst}{}
\floatname{listing}{Listing}
\title{RealAC: A Domain-Agnostic Framework for Realistic and Actionable Counterfactual Explanations}
\author{
    Asiful Arefeen, Shovito Barua Soumma, Hassan Ghasemzadeh
}
\affiliations{
    Arizona State University\\

}

\usepackage{bibentry}

\begin{document}

\maketitle

\begin{abstract}
Counterfactual explanations provide human-understandable reasoning for AI-made decisions by describing minimal changes to input features that would alter a model's prediction. To be truly useful in practice, such explanations must be realistic and feasible—they should respect both the underlying data distribution and user-defined feasibility constraints. Existing approaches often enforce inter-feature dependencies through rigid, hand-crafted constraints or domain-specific knowledge, which limits their generalizability and ability to capture complex, non-linear relations inherent in data. Moreover, they rarely accommodate user-specified preferences and suggest explanations that are causally implausible or infeasible to act upon. We introduce \textbf{\textit{RealAC}}, a domain-agnostic framework for generating realistic and actionable counterfactuals. RealAC automatically preserves complex inter-feature dependencies without relying on explicit domain knowledge—by aligning the joint distributions of feature pairs between factual and counterfactual instances. The framework also allows end-users to ``freeze'' attributes they cannot or do not wish to change by suppressing change in frozen features during optimization. Evaluations on three synthetic and two real datasets demonstrate that RealAC balances realism with actionability. Our method outperforms state-of-the-art baselines and Large Language Model-based counterfactual generation techniques in causal edge score, dependency preservation score, and IM1 realism metric and offers a solution for causality-aware and user-centric counterfactual generation. Code: \href{https://github.com/Arefeen06088/RealAC}{\textcolor{blue}{github.com/Arefeen06088/RealAC}}
\end{abstract}



\section{Introduction}

While black-box machine learning models are increasingly adopted to support decision-making in high-stakes domains, there is a growing demand for methods that can explain and justify their predictions to end-users \cite{Wachter2017CounterfactualEW}. Counterfactual explanations (CFs) have emerged as a powerful class of local interpretability techniques that respond to this need by answering “what-if” questions. CF identifies a minimal set of feature changes to an input instance that would flip the model’s prediction to a more favorable outcome. For example, a wearable-based ML model predicts that a patient is at high risk for an anxiety episode in the next hour. A CF might suggest: \textit{“Your risk would have been lower if your screen time in the past 2 hours had been under 30 minutes.”} Such explanations not only enhance transparency but also offer actionable intervention for improving future outcomes \cite{VanNostrand2024ActionableRF}. When compared against feature attribution methods like SHAP \cite{Lundberg2017AUA} or LIME \cite{Ribeiro2016WhySI}, which describe the contribution of features to a prediction, CFs offer more precise, granular, and causal insights by suggesting plausible changes that could have altered the outcome, which positions them on the third rung of Pearl’s causal hierarchy \cite{Frappier2018TheBO}. 


CFs are valuable for actionable insights, however, they are only trustworthy when they are also realistic: their feature combinations are plausible and conform to the original data’s causal structures, correlations, and distributional patterns. Earlier efforts ensured that CFs lie within the support of the training data \cite{Nemirovsky2022CounteRGANGC}. Nonetheless, realism is not only about data likelihood; it also involves preserving the causal relationships between features \cite{Mahajan2019PreservingCC}. Researchers have refined this definition over time to include consistency with causal relations and inter-feature dependencies observed in the original data \cite{Mahajan2019PreservingCC, Crupi2021CounterfactualEA}. In digital health, suggesting more steps without increased distance walked is unrealistic—even if it changes the model's prediction. Although certain methods capture linear and highly correlated dependencies between features \cite{Mahajan2019PreservingCC, Crupi2021CounterfactualEA, Xiang2022RealisticCE, Artelt2021ConvexOF}, we often exclude such features during model training to reduce redundancy, assuming they carry overlapping information. Non-linear and domain-relevant feature dependencies, however, cannot be ignored: cognitive performance may follow a parabolic relation with sleep duration [e.g., too little or too much sleep both degrade performance] \cite{Wild2018DissociableEO}, mood fluctuates cyclically with time of day \cite{Golder2011DiurnalAS}, and mental well-being often displays an inverted U-shaped curve with screen time \cite{Przybylski2017}. These relationships may not show up as strong linear correlations, but they are central to generating meaningful and feasible CFs. Prior works have attempted to preserve such structures by relying on partial causal graphs \cite{Karimi2020AlgorithmicRF}, domain knowledge and/or conditional generative models \cite{Mahajan2019PreservingCC}. Yet, these methods are often limited by their reliance on explicit structural assumptions or access to expert-provided causal diagrams, which are rarely available in high-dimensional or noisy domains like medicine. Therefore, there is a need for CF frameworks that automatically preserve complex inter-feature dependencies, including causal and non-linear interactions, without relying on domain-specific knowledge.

Beyond realism, actionability determines if a CF is implementable in practice. CFs must respect user-defined constraints—features that users cannot change or prefer not to change due to personal, physical, financial, or contextual limitations. Ignoring such local feasibility constraints undermines trust in AI and leads to inapplicable suggestions. To address actionability, prior works  incorporated user preferences via hand-coded constraints \cite{Ustun2018ActionableRI}, weighted perturbations \cite{Afrabandpey2022FeasibleAD}, randomized feature orders for prioritization \cite{Arefeen2023DesigningUB} or by assigning individual weights \cite{Arefeen2025GlyTwinDT}.

To address the dual objectives: realism and actionability, RealAC is designed as a domain-agnostic CF framework that preserves inter-feature dependencies and adheres to user-specified local feasibility constraints. RealAC is distinctive in several ways:
\begin{itemize}
    \item Prior methods often rely on fixed priors to encode dependencies during optimization \cite{Mahajan2019PreservingCC}, assume access to explicit causal graphs or structural equations \cite{Karimi2019ModelAgnosticCE}, or utilize models such as DAG-GNN to learn dependencies before perturbing latent representations \cite{Xiang2022RealisticCE}. However, the extent to which these approaches capture complex, non-linear inter-feature interactions remains unclear (Figure \ref{xiantao}). RealAC, on the other hand, minimizes the divergence in mutual information between pairs of features in the CFs and their counterparts in the original data, ensuring statistical and structural consistency in a domain-agnostic manner.

\begin{figure}[!h]
\centering
\includegraphics[width=0.8\linewidth]{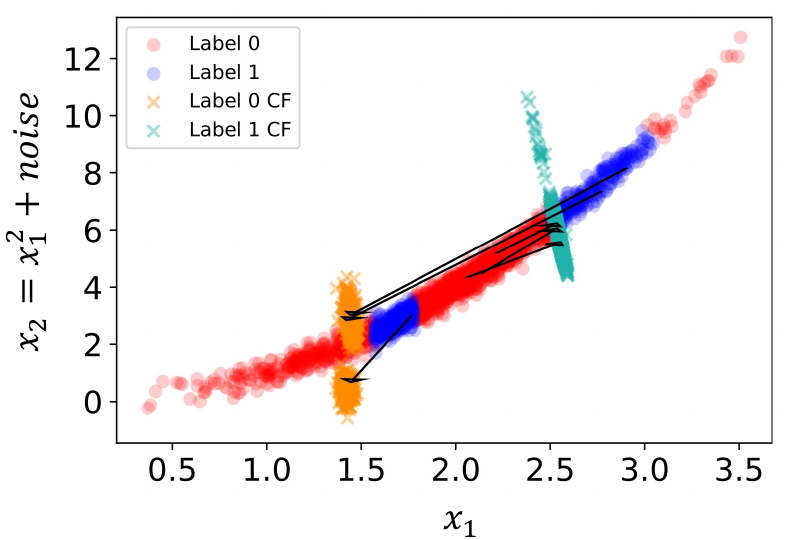}
\caption{CFs from Xiang et al. \cite{Xiang2022RealisticCE} cannot preserve feature dependency and exhibit poor diversity.}
\label{xiantao}
\end{figure}

    \item Previous work addressed user preferences either by defining local infeasibility levels \cite{Afrabandpey2022FeasibleAD} or ranking feature perturbations \cite{Arefeen2023DesigningUB}. Others, like \cite{Crupi2021CounterfactualEA}, introduced soft interventions and post-hoc feasibility adjustments via SCM-derived residuals. In contrast, RealAC embeds local feasibility constraints directly into the optimization objective using binary masking mechanism and suppressing change in immutable features. RealAC ensures features- that are immutable by nature and immutable contextually by choice- remain fixed throughout the CF generation process.

\end{itemize}
Our contributions are threefold:
\begin{itemize}
    \item[1.] We propose a novel mechanism that preserves inter-feature dependencies by minimizing the divergence in mutual information across all feature pairs between CFs and the original data. CFs generated this way align with the joint distribution of the training data—without the requirement of explicit structural assumptions or domain knowledge.

    \item[2.] RealAC integrates user-specified feasibility directly into its optimization objective through a binary masking strategy and enables both structurally immutable and contextually immutable features to remain fixed in the generated CFs.

    \item[3.] We introduce the Dependency Preservation Score (DPS) as a quantitative measure of how well inter-feature dependencies are met in generated CFs. Using three synthetic, two real datasets and including evaluations with large language models, we demonstrate that RealAC consistently outperforms existing baselines in multiple metrics.
\end{itemize}

\section{Problem Formulation}
Let \( \mathbf{x} \in \mathbb{R}^d \) be a \( d \)-dimensional feature vector representing an input instance, where each element \( x_i \) corresponds to a specific feature. Let \( \mathbf{x}^0 \) denote the original observed instance and \( \mathbf{x}^{cf} \) its CF counterpart, representing a modified input that changes the model's prediction. 

Let \( f: \mathbb{R}^d \rightarrow \mathcal{Y} \) be the black-box model under explanation, where \( f(\mathbf{x}) \) is the predicted label for input \( \mathbf{x} \), and \( \mathcal{Y} \) is the label space. The input distribution is denoted by \( D \), from which \( \mathbf{x}^0 \sim D \). \( \mathcal{F}_{fixed} \subseteq \{1, \dots, d\} \) is the set of indices corresponding to immutable features (e.g., due to physical, preferential or contextual constraints). \( \mathbf{m} \in \{0, 1\}^d \) is a binary mask vector where \( m_i = 1 \) if feature \( x_i \) is immutable (i.e., \( i \in \mathcal{F}_{fixed} \)) or non-actionable, and \( m_i = 0 \) otherwise. \( \rho(x_i, x_j) \) is a measure of dependency between features \( x_i \) and \( x_j \) which can capture the extent of inter-feature relationships. Given this setup, we pose the following research question: Can we generate a CF instance $\mathbf{x}^{cf}$ such that:
\begin{enumerate}
    \item[(1)] $\mathbf{x}^{cf}$ remains realistic in terms of inter-feature dependencies observed in the training data;
    \item[(2)] the transformation from $\mathbf{x}$ to $\mathbf{x}^{cf}$ obeys user-specified local feasibility constraints;
    \item[(3)] and, all other basic requirements of CFs like prediction flip ($f(\mathbf{x}^{cf}) = y' \neq y$) and close proximity are satisfied?
\end{enumerate}

So, the RealAC problem can be viewed as a constrained optimization problem:
\begin{align*}
\text{Find }&   \mathbf{x}^{cf} \in \mathbb{R}^d  \\
\text{such that: }\\
    f(\mathbf{x}^{cf}) = & y^{cf} \neq f(\mathbf{x}^0) \tag{Prediction Flip} \\
    \|\mathbf{x}^{cf} - \mathbf{x}^0&\|_p \text{ is minimized}, \quad p \in \{1, 2\} \tag{Proximity} \\
    \rho(x_i^{cf}, x_j^{cf}&) \approx \rho(x_i, x_j), \quad \forall (i, j) \in \mathcal{P},\ \mathcal{P} \subseteq [d] \times [d] \tag{Dependency Preservation} \\
    x_j^{cf} = x^0_j, &\quad \forall j \in \mathcal{F}_{\text{fixed}} \tag{Constraint feasibility}
\end{align*}

\section{Methodology}
In this section, we will go over the individual components built to satisfy the aforementioned constraints towards generating realistic and actionable CFs.

\subsection{Label Flip}
To ensure CF samples change the model's prediction, we begin with a label flipping constraint during optimization. Given the target class label \( y^{cf} \neq f(\mathbf{x}^0) \), we define a classification loss that encourages \( f(\mathbf{x}^{cf}) \) to align with \( y^{cf} \). The label flipping objective is formalized using a standard cross-entropy loss:
\[
\mathcal{L}_{\text{flip}} = \frac{1}{N} \sum_{n=1}^N \text{CE}(y^{cf}_n, f(\mathbf{x}^{cf}_n)),
\]
where \( N \) is the batch size and \( \text{CE} \) denotes the categorical cross-entropy loss between the desired target label \( y^{cf} \) and the model's prediction. Optimizing $\mathcal{L}_{\text{flip}}$ guides the CF generation process toward regions in feature space where the classifier assigns high confidence to the desired target class.

\subsection{Feature Dependency}

A fundamental requirement for generating realistic and trustworthy CFs is preserving inter-feature dependencies. In certain datasets collected from physical, biological, or behavioral systems, features exhibit strong pairwise dependencies due to underlying causal, physical, or semantic constraints. Ignoring these dependencies during CF generation may result in impaired interpretation and violate feasibility in downstream decision-making.

To address this, RealAC enforces the joint probability structure among feature pairs using a sampling-based approach grounded in empirical estimation.

We propose a pairwise regularization scheme $\mathcal{L}_{\text{dep}}$ based on empirical joint distribution matching to capture and preserve feature dependencies. The goal is to ensure that the marginal and joint statistics of feature pairs in $\mathbf{x}^{cf}$ closely reflect those in $\mathbf{x}^0$. For each pair of features $(i, j)$, we estimate their joint distributions in form of mutual information both in the original and CF instances as $\rho(x_i, x_j)$ and $\rho(x^{cf}_i, x^{cf}_j)$, respectively.

The dependency between feature pairs \( (x_i, x_j) \) is quantified using mutual information, defined as:
\[
\rho(x_i, x_j) = \sum_{x_i, x_j} P(x_i, x_j) \log \frac{P(x_i, x_j)}{P(x_i) P(x_j)},
\]
where \( P(x_i, x_j) \) is the joint probability, and \( P(x_i) \) and \( P(x_j) \) are the marginal probabilities. To impose this dependency on CFs, a downstream optimizer minimizes the difference between the dependency of the original feature pairs $\rho(x_i, x_j)$ and the generated pairs $\rho(x^{cf}_i, x^{cf}_j)$. This is formulated as a loss term:
\[
\mathcal{L}_{\text{dep}} = \frac{1}{|\mathcal{P}|} \sum_{(i, j) \in \mathcal{P}} \left| \rho(x_i^{cf}, x_j^{cf}) - \rho(x_i, x_j) \right|,
\]
where \( \mathcal{P} \) is the set of all continuous feature pairs, and the loss is weighted by a hyperparameter \( \lambda_{dep} \) to balance its impact.

Since estimating joint distributions directly in continuous space is computationally expensive for high-dimensional or large datasets, we discretize each feature into $B$ bins. For each feature pair $(i, j)$, the empirical joint distributions are estimated via normalized co-occurrence histograms:
\[
\hat{P}_{pq}(x_i, x_j) = \frac{1}{N} \sum_{n=1}^{N} \mathds{1}_{[x_{n,i} \in b_p]} \cdot \mathds{1}_{[x_{n,j} \in b_q]},
\]
where $N$ is the total number of samples in the bin, \( \mathds{1}_{[\cdot]} \) is the indicator function, \( b_p \) and \( b_q \) denote the \( p^{\text{th}} \) and \( q^{\text{th}} \) bins of features \( x_i \) and \( x_j \), respectively, and \( x_{n,i} \) is the \( i^{\text{th}} \) feature value of the \( n^{\text{th}} \) data sample. 

This binning strategy allows joint estimation for both continuous and categorical features. The number of bins \( B \) is selected based on sample size and feature resolution to balance granularity and statistical stability.

Using pairwise joint distributions offers a tractable approximation to the full joint and captures nuances of feature interdependence. Unlike methods that impose structure via explicit domain knowledge (e.g., structural causal models), our empirical approach is domain-agnostic, adapts to diverse tabular datasets with unknown
or heterogeneous relationships and still encourages structural consistency. Importantly, this regularizer is differentiable with respect to the perturbation and can be integrated into any CF optimization framework.

\subsection{Actionability}
Another key requirement for practically useful CF interventions is actionability — the notion that any changes suggested by the CF instance must be feasible, ethical, and aligned with user constraints. Real-world decisions often involve variables that are immutable (e.g., ethnicity), hard-to-modify (e.g., blood type, genetic predispositions), or contextually sensitive (e.g., marital status). Violating actionability constraints leads to unrealistic or unethical recommendations with limited applicability in high-stakes domains.

In RealAC, user-preferences are preserved within the optimization process using a flexible binary masking mechanism. We define a binary \textbf{actionability mask} $\mathbf{m} \in \{0, 1\}^d$ for each sample, where:
\begin{itemize}
  \item $m_j = 1$ indicates that feature $j$ is immutable or user-restricted,
  \item $m_j = 0$ implies that feature $j$ is mutable.
\end{itemize}

During optimization, we ensure that only mutable features are allowed to change from their factual values, while change in immutable ones are suppressed. This is achieved by penalizing changes to the immutable features using regularization terms for both continuous and categorical features.

\begin{itemize}
    \item For continuous features:
    \[
    \mathcal{L}_{\text{fixed, cont}} = \frac{1}{N} \sum_{n=1}^N \sum_{i \in \mathcal{C}} m_{n,i} \left(x_{n,i} - \hat{x}_{n,i}\right)^2
    \]
    where \( \mathcal{C} \) is the set of continuous features, \( N \) is the batch size, \( m_{n,i} \in \{0, 1\} \) is a mask indicating whether feature \( i \) of sample \( n \) is fixed, \( x_{n,i} \) is the factual value, and \( \hat{x}_{n,i} \) is the reconstructed or optimized value.

    \item For categorical features:
    \[
    \mathcal{L}_{\text{fixed, cat}} = \frac{1}{N} \sum_{n=1}^N \sum_{i \in \mathcal{D}} m_{n,i} \cdot \text{CE}(x_{n,i}, \hat{p}_{n,i})
    \]
    where \( \mathcal{D} \) is the set of categorical features, and \( \text{CE} \) denotes the sparse categorical cross-entropy between the factual category \( x_{n,i} \in \{0, 1, \dots, K_i-1\} \) and the predicted class probabilities \( \hat{p}_{n,i} \in \mathbb{R}^{K_i} \) output by the model.
\end{itemize}

During optimization, perturbations are applied only to mutable features. The actionability constraint is enforced via elementwise masking to merge immutable factual features with the optimizer’s updates:
\[
\mathbf{x}^{cf} = \mathbf{m} \odot \mathbf{x}^0 + (1 - \mathbf{m}) \odot \hat{\mathbf{x}}
\]
where $\odot$ denotes the Hadamard (element-wise) product and $\hat{\mathbf{x}}$ is the reconstructed value out of the optimizer.

Zeroing out gradients on immutable features prevents the optimization from altering any features deemed unchangeable. Furthermore, our framework supports dynamic reconfiguration of the actionability mask $\mathbf{m}$ and end-users to specify feature-level constraints on the fly.

This masking technique is compatible with gradient-based CF generators. Unlike prior approaches that require external penalty terms or post-hoc filtering, we incorporate actionability directly into the optimization for a constraint-aware CF generation.

\subsection{Proximity}
CFs should remain in close proximity to the factual instances. We encourage proximity by penalizing large changes to the mutable features using distance-based regularization. For the mutable continuous features, as defined by the binary mask vector $\mathbf{m}$, we use a mean squared error penalty between the factual instance and the CF:
\[
\mathcal{L}_{\text{dist, cont}} = \frac{1}{N} \sum_{n=1}^N \sum_{i \in \mathcal{C}} (1-m_{n,i}) \left(x_{n,i} - \hat{x}_{n,i}\right)^2
\]
Similarly, For categorical features, we use a cross-entropy loss between the original one-hot encoded value \( x_{n,i} \) and the predicted probability distribution \( \hat{p}_{n,i} \) over categories:
\[
\mathcal{L}_{\text{dist, cat}} = \frac{1}{N} \sum_{n=1}^N \sum_{i \in \mathcal{D}} (1-m_{n,i}) \cdot \text{CE}(x_{n,i}, \hat{p}_{n,i})
\]

$ \lambda_{fixed} > \lambda_{mse} $ should theoretically suppress change in immutable features and allow controlled mutation in the rest.
\subsection{Base Optimizer}
To optimize all aforementioned desiderata: label flipping, realism via joint distribution alignment, actionability, and proximity—RealAC adopts a variational autoencoder (VAE)-based architecture, maintaining similarity with prior works \cite{Mahajan2019PreservingCC, Panagiotou2024TABCFCE, Guyomard2022VCNetAS}. The KL divergence term, inherent to the VAE architecture through evidence lower bound (ELBO) \cite{Pawelczyk2019LearningMC}, ensures that encoder $\mathbf{q}_\phi (\mathbf{z}|\mathbf{x}^0, y^{cf})$ maps an input instance $\mathbf{x}^0$ and its target class label $y^{cf}$ to a suitable latent representation $\mathbf{z}$ and the decoder generates a CF sample $\mathbf{x}^{cf}$ from $\mathbf{z}$ and $y^{cf}$ by increasing the conditional likelihood $\mathbf{p}_\theta (\mathbf{x}^{cf}|\mathbf{z}, y^{cf})$. During optimization, we minimize a total loss composed of multiple objectives:
\begin{align*}
\mathcal{L}_{\text{total}} = &\mathcal{\lambda}_{\text{flip}}\cdot\mathcal{L}_{\text{flip}}+ \mathcal{\lambda}_{\text{dep}}\cdot\mathcal{L}_{\text{dep}}+\\ &\mathcal{\lambda}_{\text{fixed}}\cdot(\mathcal{L}_{\text{fixed,cont}}+\mathcal{L}_{\text{fixed,cat}})+\\
&\mathcal{\lambda}_{\text{dist}}\cdot(\mathcal{L}_{\text{dist, cont}}+\mathcal{L}_{\text{dist, cat}})+\\ 
& \mathcal{\lambda}_{\text{KL}}\cdot \text{KL}(\mathbf{q}_\phi (\mathbf{z}|\mathbf{x}^0, y^{cf})||\mathbf{x}^{cf}|\mathbf{z}, y^{cf})
\end{align*}
Each regularization term is modulated by a tunable weight ($\lambda_{\text{flip}}$, $\lambda_{\text{dep}}$, $\lambda_{\text{fixed}}$, $\lambda_{\text{dist}}$, $\lambda_{\text{KL}}$) to have flexible control over the impact of different constraints.

\begin{figure}[!t]
\centering
    \includegraphics[width=0.9\linewidth]{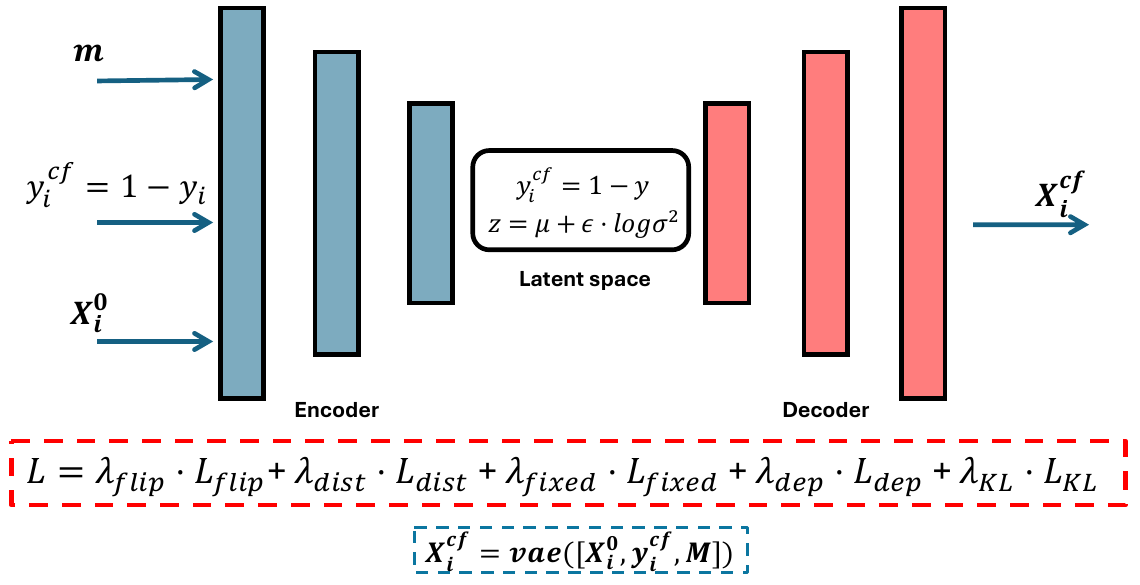}
\caption{VAE framework in RealAC receives feasibility mask, target class and factual samples as input and optimizes flip, distance, fixed feature, and dependency regularization to generate the CF.}
\label{vae_arch}
\end{figure}

\section{Experiment}
\subsubsection{\underline{Datasets}-} RealAC is tested on three synthetic and two real datasets. In \textbf{Synthetic 1} and \textbf{Synthetic 2} \cite{Xiang2022RealisticCE}, complex inter-feature relations- like sinusoidal, quadratic, exponential and linear dependencies- are simulated for a comprehensive evaluation. \textbf{Sangiovese} \cite{Mahajan2019PreservingCC} is another synthetic data that exhibits approx. linear trends among feature pairs. The two real datasets, \textbf{Diabetes} \cite{Mahajan2019PreservingCC} and \textbf{Adult} \cite{Virgolin2022OnTR} contain somewhat linear trends among some feature pairs. While Adult dataset includes both continuous and categorical features, others consist solely of continuous variables. A binary classification model is trained on each of the aforementioned datasets with accuracies of $96.53\%$, $87.83\%$, $83.12\%$, $83.65\%$, and $82.85\%$, respectively. For each dataset, user preference $\mathbf{m}$ is simulated randomly. More details on the datasets are provided in the Technical Appendix.
\subsubsection{\underline{Baselines}-} We compare RealAC against the following baselines:

\textbf{CEILS} \cite{Crupi2022LeveragingCR} is a wrapper that adds domain-level causal relationships into the CF generation process by perturbing a learned latent space. It is combined with DiCE to generate causal CFs. CEILS requires access to structural prior to preserve feature constraints.

\textbf{MCCE} \cite{Redelmeier2024MCCEMC} builds a generative model that captures how mutable features vary with respect to immutable attributes and model outcomes using autoregressive decision trees and samples realistic CFs through Monte Carlo sampling and filtering.

\textbf{C-CHVAE} \cite{Pawelczyk2019LearningMC} uses a VAE to learn a latent manifold of the data and then searches this space for valid and plausible CFs in close proximity of high-density regions.

\textbf{DiCE} \cite{Mothilal2019ExplainingML} generates a diverse set of CFs to maximize variability across solutions while also optimizing for proximity, and feasibility across local regions of the decision boundary.

\textbf{CFNOW} \cite{Oliveira2023AMA} is a model-agnostic method that employs a two-step search algorithm to explore the search space and generate valid and minimal CFs.

\textbf{NICE} \cite{Brughmans2021NICEAA} iteratively constructs CFs by replacing feature values with those from the nearest instance having a different prediction.


\textbf{SenseCF} \cite{Soumma2025SenseCFLC} prompts large language models (LLMs) to generate CFs as synthetic datapoints for training robust models. We implement SenseCF using OpenAI o3-mini and GPT 4o in a few-shot learning format using 4 random samples.

\subsubsection{\underline{Validation metrics}-} To evaluate the quality of CFs generated by RealAC, we use several validation metrics.

\textbf{\textit{Validity}} \cite{Hamman2023RobustCE} measures the \% of CFs that flip prediction; higher scores indicate effective interventions.

\textbf{\textit{Distance}} \cite{Karimi2020AlgorithmicRF} is the combination of $l_2$ distance normalized by feature range and Hamming distance. It quantifies how distant a CF is from the factual instance. Lower values reflect minimal changes.

\textbf{\textit{Causal Edge Score}} \cite{Mahajan2019PreservingCC} evaluates the causal plausibility of CFs by comparing how likely they are under the true causal distribution w.r.t their corresponding factuals. A value close to zero is optimal.

\textbf{\textit{Dependency Preservation Score (DPS)}} assesses how well inter-feature dependencies in the original data are preserved in the CFs—higher scores are better.

\textbf{\textit{IM1}} \cite{Looveren2019InterpretableCE} measures the ratio between the reconstruction errors of CFs using target class and original class autoencoder. Lower IM1 is better.

\textbf{\textit{Plausibility}} \cite{Guidotti2024CounterfactualEA} is the \% of CFs within the valid region of the original data. Values close to 1 are better.

All experiments were performed using a single compute node with access to 8 CPU cores, 16 GiB of RAM, and a single NVIDIA A100 GPU for hardware acceleration.

\begin{table*}[!h]
\footnotesize
\caption{Evaluating the CFs on \textbf{data with continuous features} only: RealAC outperforms others in causal edge score, dependency preservation score, and IM1 and achieves competitive scores in validity, distance and plausibility for all datasets.}
\label{cont_evaluation}
\centering
\begin{tabular}{p{0.7in}|C{0.34in}C{0.34in}C{0.4in}C{0.34in}C{0.34in}C{0.34in}|C{0.34in}C{0.34in}C{0.4in}C{0.34in}C{0.34in}C{0.34in}}
\toprule
 \multirow{2}{*}{Method} &\multicolumn{6}{c|}{\textbf{Synthetic 1}}& \multicolumn{6}{c}{\textbf{Synthetic 2}} \\
& \cellcolor{blue!18}val. $\uparrow$ & \cellcolor{red!25}dist. $\downarrow$ & \cellcolor{green!25}ces.$\approx0$& \cellcolor{yellow!25}dps. $\uparrow$ & \cellcolor{teal!25}IM1 $\downarrow$ & \cellcolor{orange!25}plau. $\uparrow$ & \cellcolor{blue!18}val. $\uparrow$ & \cellcolor{red!25}dist. $\downarrow$ & \cellcolor{green!25}ces.$\approx0$ & \cellcolor{yellow!25}dps. $\uparrow$ & \cellcolor{teal!25}IM1 $\downarrow$ & \cellcolor{orange!25}plau. $\uparrow$ \\
\midrule 
RealAC &  
0.996 & 0.296 & -6.558 & 0.719 & 0.776 & 1.000 & 1.000 & 0.109 & -5.892 &  0.709 & 0.980 & 0.987 \\
CEILS & 
0.870 & 0.298 & -12.260 & 0.587 & 0.832 & 1.000 &  0.930 & 0.327 & -10.231 & 0.668 & 1.261 & 1.000 
\\
MCCE & 
0.994 & 0.344 & -9.346 & 0.686 & 0.953 &  1.000 &  1.000 & 0.397 & -11.248 & 0.689 & 1.067 & 0.954 \\
\midrule
DiCE & 
0.996 & 0.481 & -11.300 & 0.546 & 0.893 &  1.000 & 1.000 & 0.449 & -21.730 & 0.506 & 1.789 & 1.000 
\\
CFNOW & 
0.343 & 0.425 & -27.674 & 0.663 & 0.925 & 1.000 & 0.575 & 0.545  & -16.110 & 0.614 & 1.125 & 0.432 
\\
NICE & 
0.212 & 0.331 & -17.552 & 0.635 & 0.913 &  1.000 & 0.734 & 0.515 & -17.500 & 0.711 &  1.271 & 1.000 
\\
C-CHVAE & 
0.996 &	0.481 &	-11.301 & 0.546 & 0.893 & 1.000 &  
1.000 & 0.374 &	-7.201 & 0.697 & 1.074 & 1.000 
\\
\midrule
SenseCF (o3) & 
0.824 & 0.373 & -9.223 & 0.793 & 2.330 & 1.000 & 0.522 & 0.195 & -9.166 & 0.768 & 1.684 & 1.000 
\\
SenseCF (4o) & 
0.141 & 0.133 & -4.658 & 0.771 & 1.896 & 1.000 & 0.138 & 0.097 & -8.951 & 0.728 & 1.609 & 1.000 
\\
\midrule \midrule
&\multicolumn{6}{c|}{\textbf{Diabetes}}& \multicolumn{6}{c}{\textbf{Sangiovese}} \\
&  \cellcolor{blue!18}val. $\uparrow$ & \cellcolor{red!25}dist. $\downarrow$ & \cellcolor{green!25}ces.$\approx0$ & \cellcolor{yellow!25}dps. $\uparrow$ & \cellcolor{teal!25}IM1 $\downarrow$ & \cellcolor{orange!25}plau. $\uparrow$ & \cellcolor{blue!18}val. $\uparrow$ & \cellcolor{red!25}dist. $\downarrow$ & \cellcolor{green!25}ces.$\approx0$& \cellcolor{yellow!25}dps. $\uparrow$ & \cellcolor{teal!25}IM1 $\downarrow$ & \cellcolor{orange!25}plau. $\uparrow$ \\
\midrule 
RealAC & 
0.987 & 2.641 & 0.278 & 0.598 & 1.476 &  0.982 & 1.000 & 0.338 & -0.600 & 0.438 & 0.667 & 0.998 
\\
CEILS & 
0.877 & 2.211 & 0.159 & 0.516 & 1.834 & 1.000 & 
0.773 &	0.417 &	-2.131 & 0.376 & 0.756 & 1.000
\\
MCCE & 
0.938 & 1.834 & 0.388 & 0.489 & 1.685 & 0.973 & 
1.000 & 0.322 & -1.127 & 0.411 & 0.693 & 1.000 
\\
\midrule
DiCE & 
0.950 & 2.311 & -2.111 & 0.473 & 3.277 & 1.000 & 1.000 & 0.492 & -0.981 & 0.425 & 0.896 & 1.000 
\\
CFNOW & 
1.000 & 2.437 & -0.431 & 0.527 & 2.433 & 0.273 & 0.598  & 0.694 & -1.969 & 0.413 & 0.9182 & 0.363 
\\
NICE & 
1.000 & 2.538 & -0.383 & 0.525 &  1.818 & 1.000 & 0.471 &  0.239 & -0.932 & 0.488 & 0.716 & 1.000 
\\
C-CHVAE & 
0.815 &	1.476 &	0.297 &	0.588 &	1.751 &	1.000  & 
0.974 &	0.471 &	-0.947 & 0.417 & 0.689 & 1.000 
\\
\midrule
SenseCF (o3) & 
0.487 & 0.596 & -0.064 & 0.543 & 2.234 & 1.000 & 0.468 & 0.168 & 0.019 & 0.546 & 0.938 & 1.000 
\\
SenseCF (4o) & 
0.558 & 0.597 & -0.089 & 0.568 & 6.008 & 0.981 & 0.311 & 0.301 & -0.017 & 0.513 & 0.868 & 0.986 
\\
\bottomrule
\end{tabular}
\end{table*}

\begin{table}[!h]
\scriptsize
\caption{Validation on \textbf{Adult data with continuous and categorical features}: RealAC performs better in validity, causal edge score, dependency preservation score, and IM1.}
\label{mixed_evaluation}
\centering
\begin{tabular}{p{0.605in}|C{0.22in}C{0.25in}C{0.345in}C{0.24in}C{0.248in}C{0.27in}}
\toprule
 \multirow{2}{*}{Method} &\multicolumn{6}{c}{\textbf{Adult}} \\
& \cellcolor{blue!18}val. $\uparrow$ & \cellcolor{red!25}dist. $\downarrow$ & \cellcolor{green!25}ces. $\approx0$ & \cellcolor{yellow!25}dps. $\uparrow$ & \cellcolor{teal!25}IM1 $\downarrow$ & \cellcolor{orange!25}plau. $\uparrow$ \\
\midrule 
RealAC &  
0.981 & 1.605 & 0.400 & 0.838 & 0.924 & 0.985 \\
CEILS & 
0.925 &	0.981 &	-3.765 & 0.739 & 1.544 & 1.000 \\
MCCE & 
0.980 & 0.484 & -4.65 & 0.642 & 1.002 & 0.937 \\
\midrule
DiCE & 
0.994 &	0.724 &	-0.480 & 0.555 & 0.956 & 0.952 \\
CFNOW & 
0.780 &	0.956 &	-0.970 & 0.435 & 1.883 & 0.810 \\
NICE & 
0.800 &	0.517 &	-0.394 & 0.573 & 1.226 & 0.926 \\
C-CHVAE & 
0.921 &	1.554 &	0.531 &	0.773 &	0.983 &	0.985  \\
\midrule
SenseCF (o3) & 
0.001 &	0.058 &	-0.007 & 0.576 & 2.516 & 0.999 \\
SenseCF (4o) & 
0.238 &	0.297 &	0.238 &	0.673 &	3.624 &	0.990 \\
\bottomrule
\end{tabular}
\end{table}

\begin{figure*}[!t]
    \centering

    \begin{subfigure}[b]{0.48\textwidth}
        \centering
        \includegraphics[width=0.495\textwidth]{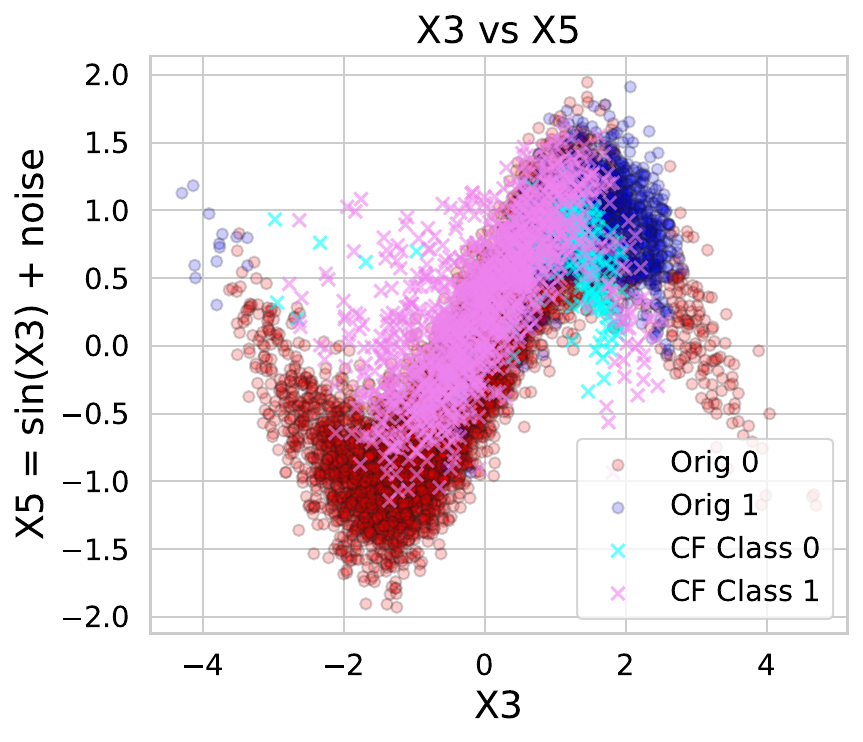}
        \includegraphics[width=0.495\textwidth]{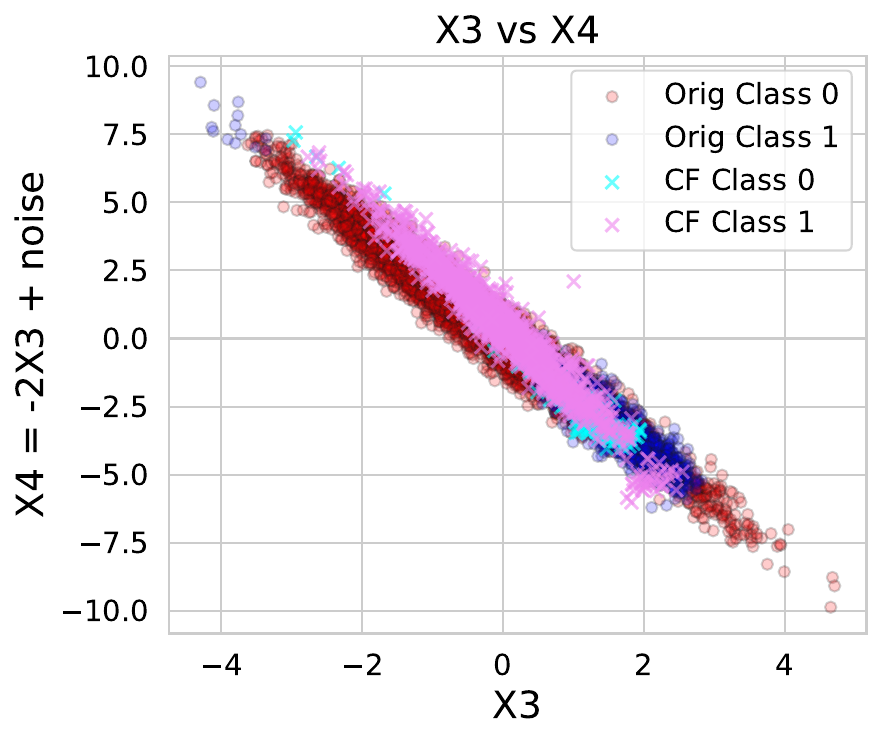}
        \caption{Synthetic 1 dataset with sinusoidal and linear dependencies}
        \label{synthetic1_2_figure}
    \end{subfigure}
    \hfill
    \begin{subfigure}[b]{0.48\textwidth}
        \centering
        \includegraphics[width=0.495\textwidth]{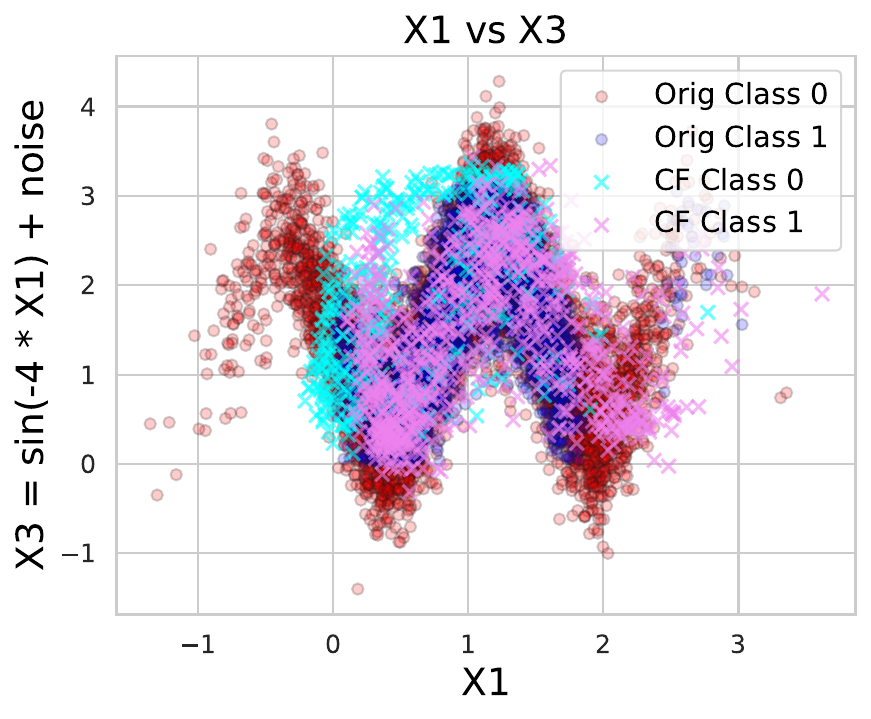}
        \includegraphics[width=0.495\textwidth]{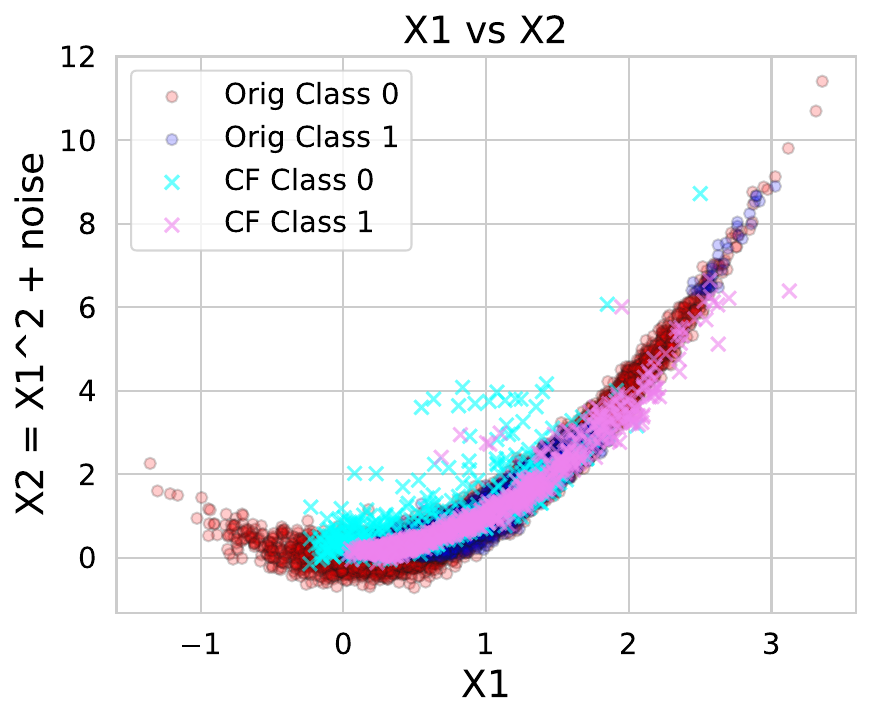}
        \caption{Synthetic 2 dataset with sinusoidal and quadratic dependencies}
        \label{synthetic2_2_figure}
    \end{subfigure}

    \vspace{0.4cm}

    \begin{subfigure}[b]{0.5\textwidth}
        \centering
        \includegraphics[width=0.495\textwidth]{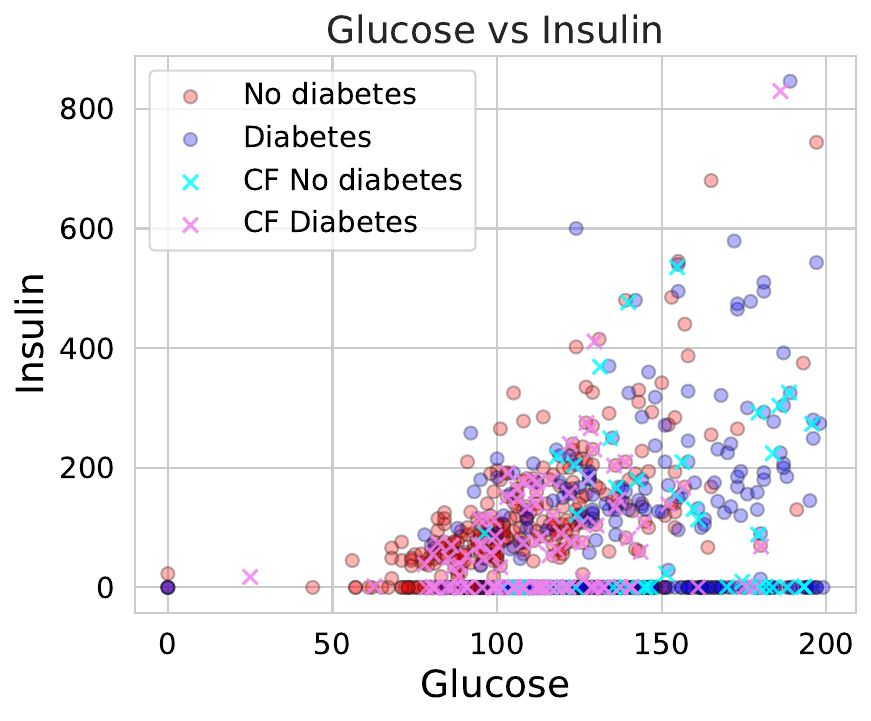}
        \includegraphics[width=0.495\textwidth]{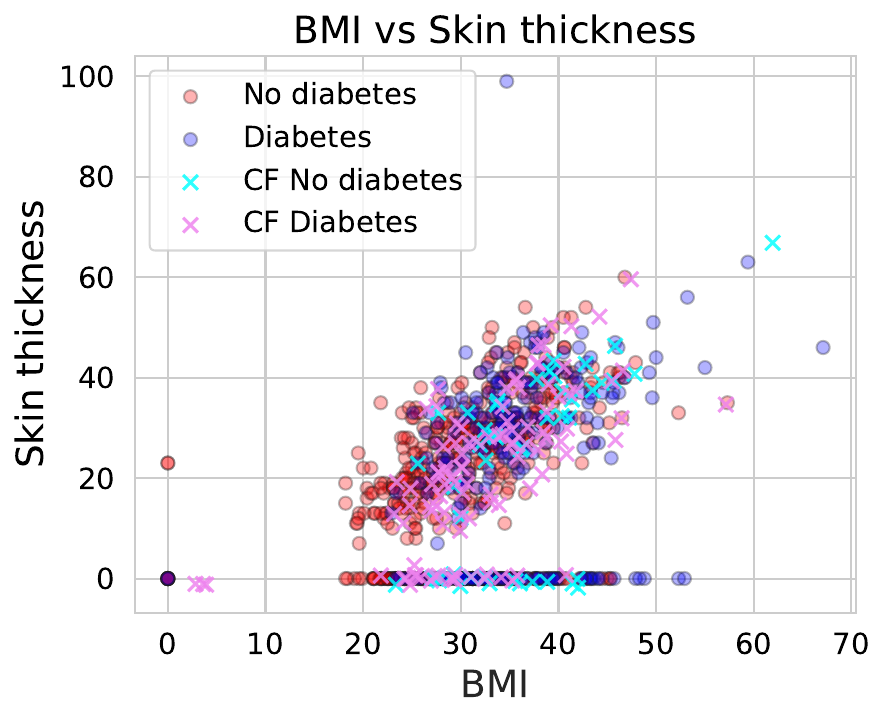}
        \caption{Diabetes dataset with somewhat linear dependencies}
        \label{diabetes_2_figure}
    \end{subfigure}
    \hfill
    \begin{subfigure}[b]{0.48\textwidth}
        \centering
        \includegraphics[width=0.495\textwidth]{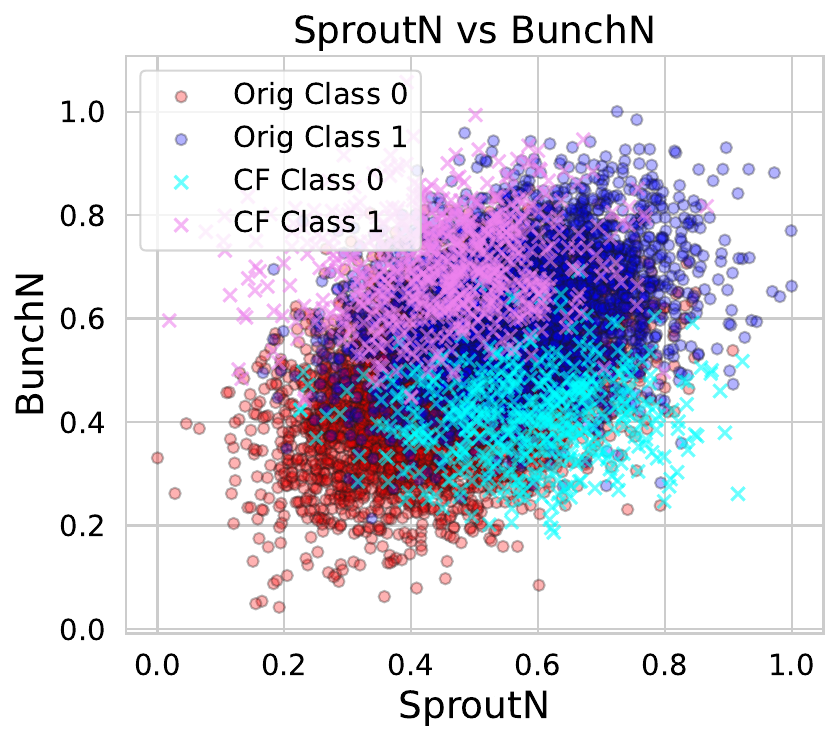}
        \includegraphics[width=0.495\textwidth]{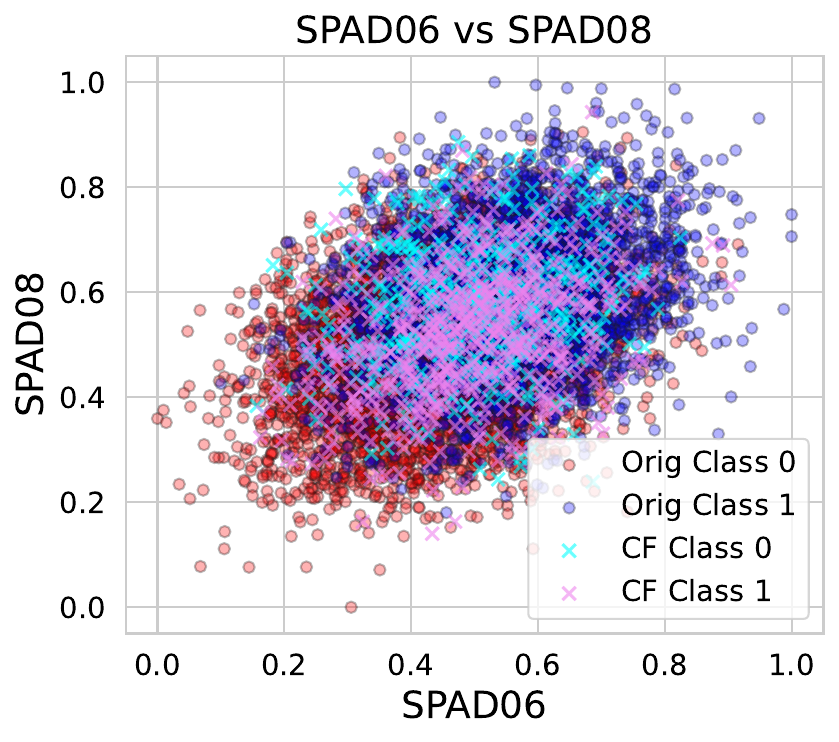}
        \caption{Sangiovese dataset with somewhat linear dependencies}
        \label{sangiovese_2_figure}
    \end{subfigure}

    \caption{Plots for visualizing complex inter-feature dependencies in different datasets and how well they are preserved in the CFs generated by RealAC.}
    \label{8_figures}
\end{figure*}

\begin{figure}[!h]
\centering
    \includegraphics[width=1\linewidth]{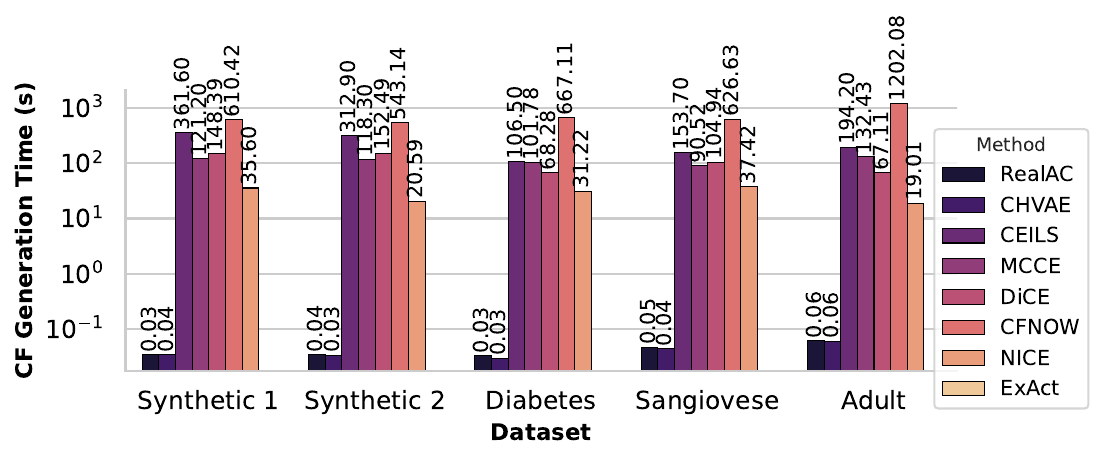}
\caption{Time to generate 100 CFs using different methods.}
\label{runtime}
\end{figure}

\section{Results}
Evaluating CF generation methods is inherently challenging \cite{DoshiVelez2017TowardsAR}. A method that rarely flips the prediction can still score highly on metrics like distance, causal edge score, dependency preservation score, and plausibility. However, such CFs fail to serve their primary purpose—offering actionable alternative scenarios that change the outcome. As a result, it is difficult to identify a CF generation method that excels across all evaluation criteria. Tables~\ref{cont_evaluation} and~\ref{mixed_evaluation} provide complete comparisons of RealAC against state-of-the-art methods, while Figure~\ref{8_figures} visually demonstrates how well RealAC preserves inter-feature dependencies. Our method consistently balances prediction flipping with realism and structural integrity. Unlike baselines that skew toward either validity or proximity, RealAC produces plausible, model-aware, and data-faithful CFs.

Table~\ref{cont_evaluation} compares the CFs generated by RealAC against those from baseline methods using the aforementioned metrics on datasets with continuous features. While RealAC does not achieve the best score in every metric, it offers a more balanced performance overall. Specifically, RealAC excels at preserving inter-feature dependencies, as reflected by its superior performance in causal edge score, dependency preservation score, and IM1. Another key observation is that performance gap between RealAC and the baselines gets more visible on the Synthetic 1 and Synthetic 2 datasets, which contain more complex inter-feature relationships, in contrast to the other datasets that include only linear dependencies. For instance, in Synthetic 1, while DiCE and NICE maintain proximity, they severely distort feature structure (DPS$<0.64$), whereas RealAC achieves the best dependency preservation (DPS $= 0.72$). CFNOW and NICE underperform in validity due to over-prioritizing minimal changes at the expense of realism. DiCE, although achieves high validity, neglects dependency structure, and often results in implausible edits.

On simpler, more linear datasets like Diabetes and Sangiovese, RealAC still retains its advantage, though margins are narrower. Here, models like NICE and DiCE perform competitively on validity and distance, but RealAC preserves realism better and offers a more faithful reconstruction of data structure (e.g., Glucose–Insulin, BMI–Skin Thickness relationships). The SenseCF approaches based on OpenAI o3-mini and GPT 4o fail to flip the class in most cases since they do not have access to the classifier but performs relatively better in other metrics.

Table~\ref{mixed_evaluation} analyzes the performance of different methods on the Adult dataset containing both categorical and continuous features. RealAC exhibits higher distances between the factual samples and their corresponding CFs as it modifies the categorical features more often. Nevertheless, RealAC aces in validation, causal edge score, dependency preservation rate and IM1 compared to the rest of the methods.

Figure~\ref{8_figures} illustrates the ability of RealAC to preserve complex inter-feature dependencies across multiple datasets. Figures~\ref{synthetic1_2_figure} and \ref{synthetic2_2_figure} show that RealAC-generated CFs—depicted by aqua and violet crosses—align closely with the nonlinear sinusoidal, linear, and quadratic trends in synthetic datasets and demonstrate high fidelity to the underlying data-patterns. Figures~\ref{diabetes_2_figure} and \ref{sangiovese_2_figure} display similar preservation of linear and correlated structures in Diabetes and Sangiovese, where CFs maintain the distributional and relational integrity of original samples (red and blue circles). 

Figure~\ref{runtime} compares the time each method takes to generate 100 CFs. RealAC's inference time is inline with that of CHVAE and faster than all others. Therefore, RealAC is both effective and efficient in generating realistic CFs with improved causal plausibility by respecting inter-feature dependencies.

\section{Ablation studies}
\subsubsection{Impact of $\mathcal{\lambda}_{dep}$:}
\begin{figure}[!h]
\centering
    \includegraphics[width=0.98\linewidth, trim=0pt 8pt 0pt 4pt, clip]{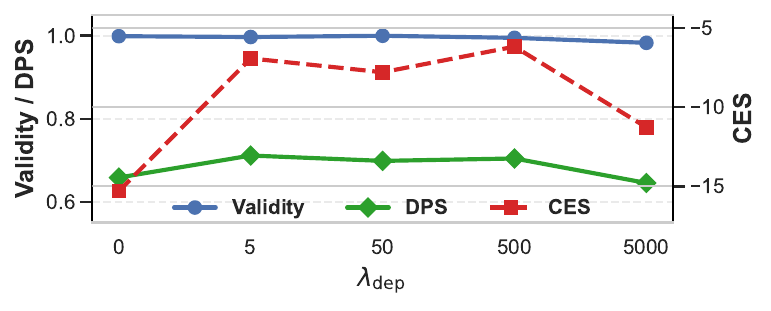}
\caption{Impact of $\mathcal{\lambda}_{dep}$ on validity, CES and DPS for Synthetic 1 dataset.}
\label{val-ces-dps_vs_lambda_dep}
\end{figure}

Figure~\ref{val-ces-dps_vs_lambda_dep} demonstrates that moderate values of $\mathcal{\lambda}_{dep}$ significantly improve the structural realism of generated CFs—reflected in higher Dependency Preservation Score and near zero Causal Edge Score—without compromising validity.
However, high $\mathcal{\lambda}_{dep}$ ($>$500) begins to hurt performance, likely because excessive emphasis on structure restricts the flexibility needed to flip predictions.

\subsubsection{Impact of number of immutable features:}
\begin{figure}[!h]
\centering
    \includegraphics[width=0.98\linewidth, trim=0pt 8pt 0pt 7pt, clip]{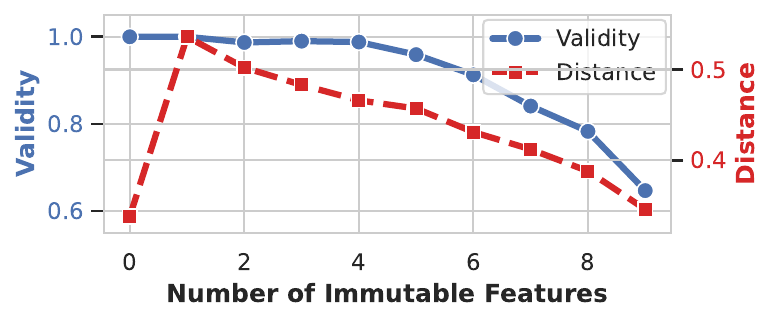}
\caption{Impact of number of immutable features on validity and distance for Sangiovese dataset.}
\label{val-dis_vs_n_fixed}
\end{figure}

Figure~\ref{val-dis_vs_n_fixed} shows that increasing the number of immutable features leads to higher distance and reduced validity, as the optimization has fewer degrees of freedom. Still, RealAC maintains high performance up to moderate constraint levels ($\leq$6 fixed features) and demonstrates robustness in user-constrained settings.

\section{Limitations}

\subsubsection{Sparsity-}Sparsity— altering the fewest possible features to generate a CF—is a desirable property of CFs, as it ensures feasibility for interventions. However, RealAC, like many methods \cite{Yang2021ModelBasedCS, Kang2020CounterfactualEB, Duong2021CausalitybasedCE}, does not explicitly optimize for sparsity. Achieving sparse CFs may require having an additional regularization term in the VAE loss function that optimizes $l_1$ distance \cite{Artelt2020ConvexDC} and penalizes the number of changed features. This could be explored in future work to enhance feasibility/actionability for end-users.

\subsubsection{Model Agnosticism-}RealAC is not model-agnostic. Like many prior methods \cite{Shao2022GradientbasedCE, Guo2021CounterNetET, Mahajan2019PreservingCC}, RealAC also requires the classifier used during CF generation to be from the family of differentiable neural networks. This restricts RealAC's applicability in settings where non-differentiable or black-box models (e.g., decision trees or ensemble methods) are used.

\subsubsection{Handling Misc. Constraints-}RealAC may struggle to capture certain domain-specific structural constraints. For example, in Synthetic 1, where a constraint like $X_6 + X_7 = $ constant holds, optimizing only for dependency preservation via joint probability-based loss $\mathcal{L}_{dep}$ is insufficient. Capturing these specific constraints requires explicit use of domain knowledge or structural priors during optimization, beyond what is learned from the data distribution alone.

\section{Conclusion}
We proposed RealAC, a GenAI-based counterfactual generation method that preserves statistical dependencies among features while satisfying infeasibility constraints specified by end users. Our approach outperforms baseline methods on multiple metrics that capture statistical coherence and causal plausibility. Despite some limitations, RealAC represents a promising step toward practical adoption of causal counterfactual explanations. As a next step, we plan to conduct a user study to evaluate its effectiveness in practical decision-making scenarios.
\newpage



\title{Technical Appendix}

\clearpage

\begin{center}
    \huge\bfseries Technical Appendix
\end{center}
\addcontentsline{toc}{part}{Technical Appendix}

\section{Dataset Description}
\subsection{Synthetic 1}
Synthetic 1 is a simulated binary classification dataset with 10,000 samples, 7 features (\texttt{X1}--\texttt{X7}) and a binary target variable \texttt{label}, using a structural equation model (SEM) that covers both linear and nonlinear dependencies.

We first sample latent variables from Gaussian distributions:
\begin{itemize}
    \item $U_1, U_2, U_3, U_4 \sim \mathcal{N}(0, 0.5)$,
    \item $U_5 \sim \mathcal{N}(0, 0.3)$,
    \item $U_6 \sim \mathcal{N}(50, 10)$.
\end{itemize}

The observable features are then defined as:
\begin{align*}
    X_1 &= U_1, \\
    X_2 &= U_2, \\
    X_3 &= 2X_1 - X_2 + U_3, \\
    X_4 &= -2X_3 + U_4, \\
    X_5 &= \sin(X_3) + U_5, \\
    X_6 &= U_6, \\
    X_7 &= 100 - U_6.
\end{align*}

The binary target label is determined based on the nonlinearity of the first five features:
\begin{itemize}
    \item Compute $\sin(X_1), \ldots, \sin(X_5)$.
    \item If more than two of these values exceed 0.5, the label (y) is set to 1; otherwise, it is 0.
\end{itemize}

This yields a nonlinear decision boundary for classification.

\paragraph{Train-Test Split.} We randomly split the dataset into 85\% training and 15\% testing sets using \texttt{train\_test\_split} and the test set is carried out in both classifier testing and RealAC testing as well. Oversampling is done to address class imbalance.

\subsection{Synthetic 2}
Synthetic 2 has five features (\texttt{X1}--\texttt{X5}) and a binary target variable \texttt{label}, based on nonlinear transformations and composite relationships. A total of 12,000 samples are created, with 90\% allocated for training.

The features are constructed using the following equations:
\begin{align*}
    X_1 &\sim \mathcal{N}(1, 0.6), \\
    X_2 &= X_1^2 + \mathcal{N}(0, 0.25), \\
    X_3 &= 1.5 + \sin(-4X_1) + \mathcal{N}(0, 0.5), \\
    X_4 &\sim \text{clipped } \mathcal{N}(2, 1) \text{ with upper bound } 3.36, \\
    X_5 &= \frac{1}{\exp(-1.5X_4)} + 2 + \mathcal{N}(0, 2).
\end{align*}

The binary label is set to 1 only if all of the following conditions are satisfied:
\[
\sin(X_1) > 0, \quad \sin(X_2) > 0, \quad \ldots, \quad \sin(X_5) > 0.
\]
Otherwise, the label is 0.
We use 90\% of the data for training and 10\% for testing. Oversampling is done to address class imbalance.

\subsection{Diabetes} Diabetes dataset contains eight continuous variables and one binzry target variable. There exist some linear trend between \texttt{Glucose-Insulin}, \texttt{BMI-Skin thickness} and \texttt{Age-number of pregnancies} pairs. Model training is done following an 80/20 split. The task is to classify if a subject has diabetes or not.

\subsection{Sangiovese} Sangiovese is a synthetic dataset with 13 continuous features. There are some linear relations between \texttt{BunchN-SproutN} and \texttt{SPAD06-SPAD08}. The original data split is followed.

\subsection{Adult}
We use the Adult Income dataset where the task is to predict whether an individual's income exceeds \$50K per year based on demographic and employment features. From the original dataset, we drop four non-essential or high-cardinality columns: \texttt{fnlwgt}, \texttt{workclass}, \texttt{occupation}, and \texttt{native-country}. The remaining features include both numerical and categorical attributes relevant to income prediction. We randomly downsample the negative class (\texttt{income = 0}) to ensure a more balanced class distribution for training. The dataset is split into 90\% training and 10\% testing using stratified sampling.

\section{Classifier Description}
To ensure maximum compatibility across all methods, we trained a straightforward MLP binary classifier for each dataset. All hyperparameters for data-specific MLP classifier are given in Table~\ref{nn_hyperparams}.

\begin{table}[!h]
\centering
\small
\caption{Classifier architectures and hyperparameters for each dataset.}
\label{nn_hyperparams}
\renewcommand{\arraystretch}{1.3}
\begin{tabular}{@{}p{2.8cm} p{5.3cm}@{}}
\toprule
\textbf{Dataset} & \textbf{Model Configuration} \\
\midrule

\textbf{Synthetic 1} &
\textbf{Layers:} Dense(64, ReLU) $\rightarrow$ Dense(32, ReLU) $\rightarrow$ Dense(16, ReLU) $\rightarrow$ Dense(2, Sigmoid) \newline
\textbf{Optimizer:} Adam \newline
\textbf{Loss:} Binary Crossentropy \newline
\textbf{Epochs:} 100, \textbf{Batch Size:} 16 \\

\midrule
& \textbf{Acc:} 0.9653, \textbf{F1 score:} 0.8725 \\
\midrule \midrule

\textbf{Synthetic 2} &
\textbf{Layers:} Dense(32, ReLU, L1=0.01) $\rightarrow$ Dense(32, ReLU) $\rightarrow$ Dense(64, ReLU) $\rightarrow$ Dense(2, Sigmoid) \newline
\textbf{Optimizer:} Adam (lr = 5e-4) \newline
\textbf{Loss:} Binary Crossentropy \newline
\textbf{Epochs:} 800, \textbf{Batch Size:} 16 \\

\midrule
& \textbf{Acc:} 0.8783 , \textbf{F1 score:} 0.8011 \\
\midrule \midrule

\textbf{Diabetes} &
\textbf{Layers:} Dense(64, LeakyReLU, He init) $\rightarrow$ BN $\rightarrow$ Dropout(0.4) $\rightarrow$ Dense(32, LeakyReLU) $\rightarrow$ BN $\rightarrow$ Dropout(0.4) $\rightarrow$ Dense(16, LeakyReLU) $\rightarrow$ BN $\rightarrow$ Dropout(0.2) $\rightarrow$ Dense(2, Sigmoid) \newline
\textbf{Optimizer:} Adam (lr = 0.01) \newline
\textbf{Loss:} Binary Crossentropy \newline
\textbf{Epochs:} 80, \textbf{Batch Size:} 16 \\

\midrule
& \textbf{Acc:} 0.8311, \textbf{F1 score:} 0.7547 \\
\midrule \midrule

\textbf{Sangiovese} &
\textbf{Layers:} Dense(512, ReLU) $\rightarrow$ Dense(256, ReLU) $\rightarrow$ Dense(128, ReLU) $\rightarrow$ Dense(64, ReLU) $\rightarrow$ Dense(2, Softmax) \newline
\textbf{Optimizer:} Adam (lr = 0.001) \newline
\textbf{Loss:} Categorical Crossentropy\newline
\textbf{Epochs:} 60, \textbf{Batch size:} 32\\

\midrule
& \textbf{Acc:} 0.8365, \textbf{F1 score:} 0.8149\\
\midrule \midrule

\textbf{Adult} &
\textbf{Layers:} Dense(256, ELU) $\rightarrow$ Dense(128, ELU) $\rightarrow$ Dense(64, ELU) $\rightarrow$ Dense(32, ELU) $\rightarrow$ Dense(2, Sigmoid) \newline
\textbf{Optimizer:} Adam (lr = 0.001) \newline
\textbf{Loss:} Binary Crossentropy \newline
\textbf{Epochs:} 120, \textbf{Batch Size:} 16 \\
\midrule
& \textbf{Acc:} 0.8285 , \textbf{F1 score:} 0.8404 \\
\midrule 
\bottomrule
\end{tabular}
\end{table}

\section{Optimizer/VAE Description}
Details of the VAE optimizers are given in Table~\ref{vae_hyperparams}.

\begin{table}[!h]
\centering
\footnotesize
\caption{VAE optimizer hyperparameters for each dataset.}
\label{vae_hyperparams}
\renewcommand{\arraystretch}{1.3}
\begin{tabular}{@{}p{1.3cm} p{6.3cm}@{}}
\toprule
\textbf{Dataset} & \textbf{Model Configuration} \\
\midrule \midrule

\textbf{Synthetic 1} &
N SAMPLES = 10000, TEST SIZE = 0.15\newline
RANDOM STATE = 42, LATENT DIM = 15\newline
BINS = 50, LAMBDA FLIP = 1.0\newline
LAMBDA DEP = 4.0, LAMBDA SUM = 2.0, for $X_6 + X_7 = k$\newline
LAMBDA MSE = 2.0, LAMBDA KL = 0.05\newline
LAMBDA FIXED = 0.5, EPOCHS = 100\newline
BATCH SIZE = 16, LEARNING RATE = 1e-2
\\

\midrule \midrule

\textbf{Synthetic 2} &
N SAMPLES = 12000, TEST SIZE = 0.1 \newline
RANDOM STATE = 42, LATENT DIM = 10\newline
BINS = 50, LAMBDA FLIP = 555.0\newline
LAMBDA DEP = 10.0, LAMBDA MSE = 830.0\newline
LAMBDA KL = 10.0, LAMBDA FIXED = 0.5 \newline
EPOCHS = 300, BATCH SIZE = 64\newline
LEARNING RATE = 5e-4, FIXED SIZE = 0
 \\

\midrule \midrule

\textbf{Diabetes} &
N FEATURES = 8, LATENT DIM = 15 \newline
BINS = 50, LAMBDA FLIP = 10.0 \newline
LAMBDA DEP = 2.0, LAMBDA MSE = 0.6 \newline
LAMBDA KL = 0.005, LAMBDA FIXED = 500 \newline
EPOCHS = 500, BATCH SIZE = 16 \newline
LEARNING RATE = 5e-4, FIXED SIZE = 0 \newline
TEST SIZE = 0.2, RANDOM STATE = 34 \\

\midrule \midrule

\textbf{Sangiovese} &
N FEATURES = 13, LATENT DIM = 15 \newline
BINS = 50, LAMBDA FLIP = 3.0 \newline
LAMBDA MI = 10.0, LAMBDA MSE = 6.5 \newline
LAMBDA KL = 0.0001, EPOCHS = 120 \newline
BATCH SIZE = 16, LEARNING RATE = 5e-4\\

\midrule \midrule

\textbf{Adult} &
TEST SIZE = 0.1, RANDOM STATE = 42\newline
LATENT DIM = 20, BINS = 50\newline
LAMBDA FLIP = 200.0, LAMBDA DEP = 6.0\newline
LAMBDA MSE = 4.0, LAMBDA KL = 0.00001\newline
LAMBDA FIXED = 0.5, EPOCHS = 80\newline
BATCH SIZE = 16, LEARNING RATE = 1e-3\newline
N FIXED = 2
\\
\midrule 
\bottomrule
\end{tabular}
\end{table}

\subsection{Validation Metrics}

We assess the quality of generated counterfactuals (CFs) using several validation metrics. Let $x_i$ denote the $i$-th factual sample, $\tilde{x}_i$ its corresponding counterfactual, and $X$ the original dataset.

\paragraph{1. Validity}
Validity measures the fraction of generated counterfactuals that flip the predicted class label:
\[
\text{Validity} = \frac{1}{N} \sum_{i=1}^{N} \mathds{1}\left[ f(\tilde{x}_i) \neq f(x_i) \right]
\]
where $f(\cdot)$ is the prediction function.

\paragraph{2. Distance (Normalized L1)}
The average normalized L2 distance between factuals and counterfactuals is defined as:
\[
\text{Distance} = \frac{1}{N} \sum_{i=1}^{N} \sum_{j=1}^{d} \frac{ |x_{ij} - \tilde{x}_{ij}|_2 }{R_j}
\]
where $R_j = \max(x_{j}) - \min(x_{j})$ is the range of feature $j$ over the original dataset. For categorical features, we use 0 if unchanged and 1 otherwise.

\paragraph{3. Causal Edge Score}
This metric quantifies how well the counterfactuals preserve known linear causal relationships. For a known linear dependency $x_j = a x_k + b + \epsilon$, the causal edge score is defined as:
\[
\text{CES} = \frac{1}{N} \sum_{i=1}^{N} \left[ \log p(\tilde{x}_{ij} \mid \tilde{x}_{ik}) - \log p(x_{ij} \mid x_{ik}) \right]
\]
Assuming Gaussian residuals, we compute log-likelihoods under a model fit on the training data.

\paragraph{4. Dependency Preservation Score (DPS)}
This score measures how well structural feature dependencies (e.g., linear correlations) are maintained in the CFs. For each known dependency $x_j \leftarrow x_k$, we compute:
\[
\text{DPS} = \frac{1}{N} \sum_{i=1}^{N} \exp \left( -\frac{| \hat{x}_{ij} - \tilde{x}_{ij} |}{\sigma_j} \right)
\]
where $\hat{x}_{ij}$ is the predicted value of $x_j$ given $\tilde{x}_{ik}$ using a regression model trained on original data, and $\sigma_j$ is the standard deviation of $x_j$.

\paragraph{5. IM1 Score}
IM1 is a fairness-aware metric that evaluates whether counterfactuals induce unintended changes in protected attributes. Let $\mathcal{S}$ denote the set of sensitive features. Then:
\[
\text{IM1} = \frac{1}{N} \sum_{i=1}^{N} \sum_{j \in \mathcal{S}} \mathds{1}\left[ x_{ij} \neq \tilde{x}_{ij} \right]
\]

\paragraph{6. Plausibility}
Plausibility checks whether counterfactuals remain within the support of the training data:
\[
\text{Plausibility} = \frac{1}{N} \sum_{i=1}^{N} \mathds{1} \left[ \tilde{x}_i \in \mathcal{R}(X) \right]
\]
where $\mathcal{R}(X)$ denotes the feature-wise value range of the training dataset. In practice, this metric computes the percentage of CFs whose continuous feature values fall within the min-max range of the training data.

\section{Prompt Details}
To evaluate the capabilities of large language models (LLMs) in generating realistic and actionable counterfactuals, we designed structured prompts tailored to each dataset. These prompts serve to ground the LLMs with task-relevant definitions, domain constraints, causal relationships, and few-shot examples. Each prompt is modular—sharing a common format but adapted to dataset-specific features and label rules.

All prompts used to generate counterfactuals from large language models (OpenAI o3-mini and GPT-4o) follow a unified structure designed to ensure clarity, consistency, and alignment with the underlying data-generating process. Each prompt includes the following core components:
\begin{enumerate}
    \item \textbf{Role Specification:} The language model is instructed to act as a knowledgeable and precise data scientist responsible for generating realistic counterfactuals.
    \item \textbf{Definition Block:} Clear definitions of counterfactuals and realism are provided to anchor the model's objective and constraints. Realism is defined in terms of structural, statistical, and causal alignment with the original dataset.
    \item D\textbf{ataset-Specific Details:} Each prompt describes the tabular data structure, including feature names, valid ranges, and any known causal or statistical relationships (e.g., deterministic functions or transformations). The binary label computation rule is explicitly defined for classification tasks.
    \item \textbf{Few-Shot Examples (optional):} For datasets where in-context demonstrations are helpful, a few original data points and their corresponding labels are shown to guide the model. In our study we used four examples.
    \item \textbf{Task Instruction:} The model is asked to minimally modify the features necessary to flip the predicted label, while preserving data realism. The expected output format is restricted to a list of only the changed features, enclosed in custom tags (e.g.,$ <new> ... </new>$), without additional explanation or commentary.
\end{enumerate}
\textbf{Common structure:}

\begin{tcolorbox}[
    colback=yellow!10,
    colframe=red!60!black,
    width=\linewidth,
    boxrule=0.4pt,
    arc=1mm,
    left=4pt, right=4pt, top=2pt, bottom=4pt,
    enhanced,
    sharp corners,
    breakable
]
\ttfamily
\textbf{ROLE:} \\
You are a precise and knowledgeable data scientist. Your task is to generate realistic counterfactuals for structured tabular data.

\textbf{DEFINITION:} \\
- A \textbf{counterfactual} is a modified version of an original data point that changes the model’s predicted outcome. \\
- A \textbf{realistic} counterfactual must satisfy structural, statistical, and causal constraints consistent with the data-generating process.

\textbf{DATASET:} \\
- List of features with ranges and descriptions. \\
- Causal or statistical relationships between features \\
- Label generation rule (e.g., threshold logic, aggregation over features).

\textbf{FEW-SHOT EXAMPLES:} \\
Example 1: ..., ... Example 4: ... 

\textbf{TASK INSTRUCTION:}
\end{tcolorbox}

\noindent \textit{Note: The role specification and definition of counterfactuals and realism are shared across all prompts and are provided in the common structure above. Dataset-specific details, label rules, and example formatting are shown below.}

\vspace{0.5em}
\noindent Below, we provide dataset-specific prompt variants adapted from the shared structure above. Each prompt reflects the unique features, causal dependencies, and label rules of the corresponding dataset.

\subsubsection{Synthetic-1}

\begin{tcolorbox}[
    colback=gray!10,
    colframe=red!60!black,
    width=\linewidth,
    boxrule=0.4pt,
    arc=1mm,
    left=4pt, right=4pt, top=2pt, bottom=4pt,
    enhanced,
    sharp corners,
    breakable
]
\ttfamily
\textbf{ROLE:} \\
\textbf{DEFINITION:} \\


\textbf{Data structure:} \\
- The data follows a known causal structure, e.g., X3 = 2 * X1 - X2 + noise. Some variables are deterministic functions of others. Noise terms follow Gaussian distributions, e.g., $\varepsilon \sim \mathcal{N}(0, \sigma^2)$.
\vspace{1em}
Your goal is to: \\
- Generate a counterfactual that flips or maintains the label, as specified. \\
- Modify as few variables as possible. \\
- Ensure structural and statistical realism. \\
- Leave unrelated or independent variables unchanged unless strictly necessary. \\
- \textbf{Only} return numeric values for changed features. \\

Label rule: \\
- Compute $\sin(X_1)$ through $\sin(X_5)$ \\
- If more than 2 of these values exceed 0.5, then label = 1; otherwise, label = 0.

Here are a few examples: \\
Example 1: X1: \{...\}, X2: \{...\}, X3: \{...\}, ..., label: \{...\} \\
Example 2: ... \\
Example 3: ... \\
Example 4: ...\\

Now generate a realistic counterfactual by changing as few features as necessary for the following instance to flip the label: \\
Target instance: X1: \{...\}, X2: \{...\}, ..., label: \{...\} \\
Enclose the generated text within <new> tags.
\end{tcolorbox}

\subsubsection{Synthetic-2}
\begin{tcolorbox}[
    colback=green!10,
    colframe=brown!60!black,
    width=\linewidth,
    boxrule=0.4pt,
    arc=1mm,
    left=4pt, right=4pt, top=2pt, bottom=4pt,
    enhanced,
    sharp corners,
    breakable
]
\ttfamily

\textbf{ROLE:} \\
\textbf{DEFINITION:} \\

\textbf{Data structure:} \\
- The dataset follows a known causal structure, e.g.,$X_2 = X_1^2 + \varepsilon_1$ \\
- Some variables are deterministic or near-deterministic functions of others. \\
- Noise terms follow normal distributions, e.g., $\varepsilon \sim \mathcal{N}(0, \sigma^2)$.

Your goal is to: \\
- Generate a counterfactual that flips or maintains the label, as specified. \\
- Modify as few variables as possible. \\
- Ensure structural and statistical realism. \\
- Leave unrelated or independent variables unchanged unless strictly necessary. \\
- \textbf{Only} return numeric values for changed features. \\

Label rule: \\
- Compute $\sin(X_1)$ through $\sin(X_5)$ \\
- If \textbf{ALL} of these values exceed 0, then label = 1; otherwise, label = 0.

Here are a few examples: \\
Example 1: X1: \{...\}, X2: \{...\}, X3: \{...\}, ..., label: \{...\} \\
Example 2: ... \\
Example 3: ... \\
Example 4: ...\\

Now generate a realistic counterfactual by changing as few features as necessary for the following instance to flip the label: \\
Target instance: X1: \{...\}, X2: \{...\}, ..., label: \{...\}\\
Enclose the generated text within <new> tags.
\end{tcolorbox}

\newpage
\subsubsection{Diabetes}

\begin{tcolorbox}[
    colback=blue!10,
    colframe=brown!60!black,
    width=\linewidth,
    boxrule=0.4pt,
    arc=1mm,
    left=4pt, right=4pt, top=2pt, bottom=4pt,
    enhanced,
    sharp corners,
    breakable
]
\ttfamily

\textbf{ROLE:} \\
\textbf{DEFINITION:} \\

\textbf{Data structure:} \\
Each patient record consists of the following numerical features: \\
- Pregnancies: Number of times pregnant (range: 0--17) \\
- Glucose: Plasma glucose concentration at 2 hours in OGTT (0--199) \\
- BloodPressure: Diastolic blood pressure in mm Hg (0--122) \\
- SkinThickness: Triceps skin fold thickness in mm (0--99) \\
- Insulin: 2-hour serum insulin (0--846 $\mu$U/ml) \\
- BMI: Body mass index (0--67.1) \\
- DiabetesPedigreeFunction (DPF): Risk based on family history (0.078--2.42) \\
- Age: Age in years (21--81) \\
- Outcome: Binary target (0 = non-diabetic, 1 = diabetic)

Here are a few examples: \\
Example 1: Pregnancies: \{\}, Glucose: \{\}, ..., Outcome: \{\} \\
Example 2: ... \\
Example 3: ... \\
Example 4: ...\\

\textbf{Target instance:} \\
Pregnancies: \{...\}, Glucose: \{...\}, BloodPressure: \{...\}, ..., Outcome: \{instance label\}

\textbf{Task:} \\
Now generate a realistic counterfactual by changing as few features as necessary to flip the label, while preserving medical and statistical realism. Return only the changed features within <new> tags.
\end{tcolorbox}

\newpage
\subsubsection{Sangiovese}

\begin{tcolorbox}[
    colback=brown!10,
    colframe=red!60!black,
    width=\linewidth,
    boxrule=0.4pt,
    arc=1mm,
    left=4pt, right=4pt, top=2pt, bottom=4pt,
    enhanced,
    sharp corners,
    breakable
]
\ttfamily

\textbf{ROLE:} \\
\textbf{DEFINITION:} \\

\textbf{Data structure:} \\
This dataset captures physiological and chemical traits of grapevines. It includes the following features: \\
SproutN, BunchN, GrapeW, WoodW, SPAD06, NDVI06, SPAD08, NDVI08, Acid, Potass, Brix, pH, Anthoc, label\\
Structural relationships: \\
- BunchN depends linearly on SproutN \\
- SPAD08 depends linearly on SPAD06 \\
- All other features are independent or non-linearly related.\\
- Maintain structural relationships: \\
  \quad - If SproutN is changed, update BunchN accordingly \\
  \quad - If SPAD06 is changed, adjust SPAD08 to reflect the dependency \\

Here are a few examples: \\
Example 1: SproutN: \{...\}, BunchN: \{...\}, ...,  GrapeW: \{...\}, Outcome: \{\} \\
Example 2: ... \\
Example 3: ... \\
Example 4: ...\\

\textbf{Target instance:} \\
SproutN: \{...\}, BunchN: \{...\}, ..., GrapeW: \{...\}, label: \{instance label\}

\textbf{Task:} \\
Now generate a realistic counterfactual by changing as few features as necessary to flip the label, while ensuring all changes are consistent with biological and structural constraints. Return only the changed features within <new> tags.
\end{tcolorbox}

\newpage

\subsubsection{Adult}

\begin{tcolorbox}[
    colback=cyan!10,
    colframe=red!60!black,
    width=\linewidth,
    boxrule=0.4pt,
    arc=1mm,
    left=4pt, right=4pt, top=2pt, bottom=4pt,
    enhanced,
    sharp corners,
    breakable
]
\ttfamily

\textbf{ROLE:} \\
\textbf{DEFINITION:} \\

\textbf{Data structure:} \\
Each individual is described using a mix of numerical and categorical features: \\
- age: Integer (person's age) \\
- education: Categorical (e.g., Bachelors, HS-grad, 11th, etc.) \\
- educational-num: Integer representing years of education (must be consistent with education level) \\
- marital-status: Categorical (e.g., Married, Divorced, Never-married) \\
- relationship: Categorical (e.g., Wife, Own-child, Husband, Not-in-family) \\
- race: Categorical (e.g., White, Black, Asian-Pac-Islander) \\
- gender: Categorical (Male, Female) \\
- capital-gain: Non-negative integer \\
- capital-loss: Non-negative integer \\
- hours-per-week: Integer (weekly work hours) \\
- income: Binary label (0 = $\leq$50K, 1 = >50K)\\

Here are a few examples: \\
Example 1:age: \{...\}, education: \{...\}, ..., income: \{...\} \\
Example 2: ... \\
Example 3: ... \\
Example 4: ...\\

\textbf{Target instance:} \\
age: \{...\}, education: \{...\}, educational-num: \{...\}, ..., income: \{instance\_inome\}

\textbf{Task:} \\
Now generate a realistic counterfactual by changing as few features as necessary to flip the label, while preserving data realism and attribute consistency. Return only the changed features within <new> tags.
\end{tcolorbox}

\end{document}